# Isometric Immersion Learning with Riemannian Geometry

Zihao Chen    Wenyong Wang    Yu Xiang

**Abstract**

Manifold learning has been proven to be an effective method for capturing the implicitly intrinsic structure of non-Euclidean data, in which one of the primary challenges is how to maintain the distortion-free (isometry) of the data representations. Actually, there is still no manifold learning method that provides a theoretical guarantee of isometry. Inspired by Nash's isometric theorem, we introduce a new concept called isometric immersion learning based on Riemannian geometry principles. Following this concept, an unsupervised neural network-based model that simultaneously achieves metric and manifold learning is proposed by integrating Riemannian geometry priors. What's more, we theoretically derive and algorithmically implement a maximum likelihood estimation-based training method for the new model. In the simulation experiments, we compared the new model with the state-of-the-art baselines on various 3-D geometry datasets, demonstrating that the new model exhibited significantly superior performance in multiple evaluation metrics. Moreover, we applied the Riemannian metric learned from the new model to downstream prediction tasks in real-world scenarios, and the accuracy was improved by an average of 8.8%.

## Introduction

As highlighted in a study published in Nature, scientists have found that geometry-based manifold learning is instrumental in integrating scientific representations of discrete non-Euclidean data. These representations are presented as compact mathematical statements of physical relationships, prior distributions, and other complex descriptors (Wang et al. 2023). Of these, with regard to the primary physical relationship of geometric distance between data samples, extensive research has focused on identifying an appropriate set of latent space coordinates that preserve the isometric property of sample distance within the learned data manifold (Meilă et al. 2024, Choi et al. 2024). Here we take a simple example and assume that a collection of discrete data samples $X$ is observed to exist in Euclidean space $\mathbb{R}^s$. In manifold learning, we commonly seek one smooth map $\mathcal{F}: \mathbb{R}^s \to \mathcal{M}$, find a physically meaningful sub-manifold $\mathcal{M}$, $\mathcal{M} \subset \mathbb{R}^s$ from the $\mathbb{R}^s$, and map the data from $\mathbb{R}^s$ to the manifold $\mathcal{M}$ with minimal distortion. Theoretically, if a mapping preserves geometric information without distortion, then the mapping is isometric immersion (Meilă et al. 2024).

Here, we encounter a substantial gap between theoretical concepts and practical techniques. Although Nash has demonstrated the feasibility of isometric immersion (MISHACHEV et al. 2004), there is still a lack of theoretically rigorous research specifically dedicated to its implementation. In non-neural network based solutions, most current methods aim to convert data from the Euclidean space to a lower-dimensional latent space while preserving local structural invariance of the data (McInnes et al. 2018). For example, some studies proposed using a local linear embedding method to construct a manifold that ensures Euclidean representation (distance, angle, etc) invariance during the process of reducing dimensionality (Xiang et al. 2023, Xiao et al. 2023, Yuanhong et al. 2022). However, previous studies have indicated that, except in very specific instances, these approaches do not ensure isometric mapping in Riemannian space (Meilă et al. 2024). In terms of neural network based solutions, some studies have emphasized that the representation learning model in manifold space can be improved by incorporating the concept of isometric, enabling more comprehensive data analysis within the latent space (Yonghyeon et al. 2021, Kato et al. 2020). Nevertheless, these approaches lack the capacity for metric learning and rely on the assumption that the latent manifold conforms to Euclidean space.

In brief, current approaches either heavily depend on assumptions about the representations' spatial distribution or neglect to integrate geometrical knowledge into the models. There is currently an absence of a method that correctly introduces Riemannian geometric priors to model, achieving isometric immersion during the learning.

In this paper, to the best of our knowledge, we are the first to propose a neural network-based method to implement the isometric immersion theorem in Riemannian space. In general, our contributions include:

- Based on the theory of Riemannian geometry, we propose the concept of isometric immersion learning.
- We explicitly leverage a series of geometric priors as loss functions to guarantee that the novel unsupervised metric and manifold learning model theoretically follows the concepts of isometric immersion, and derive a training method for this model based on the MLE (Maximum Likelihood Estimation) perspective.
- We evaluate our model on both toy datasets and real-world datasets to demonstrate its effectiveness.

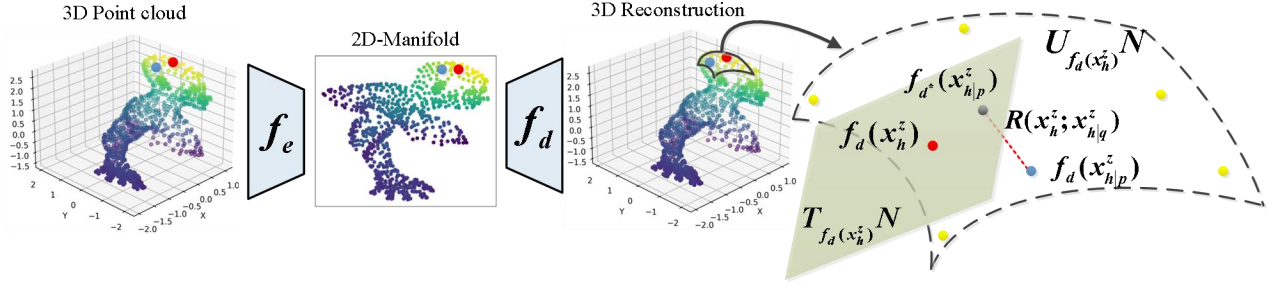

Figure 1: Use the neighborhood to approximately describe the tangent space

- We utilize the model in the field of intelligent aerodynamics. The Riemannian metric learned from the new model is applied to downstream tasks in real-world scenarios, highlighting its practicality.

## Preliminary

We provide a concise introduction to the essential definitions of Riemannian geometry and several known model properties that are utilized in this paper.

Firstly, we give the following definitions:

**Definition 1(Tangent map & pullback).** *A linear map transformation between the tangent spaces $F_*: T_p\mathcal{M} \to T_{F(p)}\mathcal{N}$ which is called the tangent map or differential of the smooth map $F$ at point $p$.*

*Conversely, $F^*: T_{F(p)}\mathcal{N} \to T_p\mathcal{M}$ is called the pullback of the smooth map $F$ at point $p$, also the dual map of $F_*$.*

**Definition 2(Immersion).** *An immersion is a differentiable map between manifolds whose tangent map is everywhere injective. Explicitly, $f: \mathcal{M} \to \mathcal{N}$ is an immersion if $F_*: T_p\mathcal{M} \to T_{F(p)}\mathcal{N}$ is an injective at every point $p$ of $\mathcal{M}$.*

Ideally, we would like a manifold learning algorithm to produce an embedding that is isometric (Dollár et al. 2007). The definition of isometric mapping is as follows:

**Definition 3(Isometric immersion).** *We set $f: (\mathcal{M}, g) \to (\mathcal{N}, h)$ as a smooth mapping between Riemannian manifolds. If $g = f^*h$, that is, for $\forall x \in \mathcal{M}$, and $\forall p, q \in T_x\mathcal{M}$, we have:*

$$h(f_*(p), f_*(q)) = g(p, q) \quad (1)$$

*where if $f$ is an immersion, then $f$ is said to be the isometric immersion from Riemannian manifold $(\mathcal{M}, g)$ to $(\mathcal{N}, h)$, and $g$ is the pullback metric. In Riemannian geometry, isometric immersion is a smooth immersion that preserves length of curves.*

With the above knowledge of Riemannian geometry, we are able to provide the Nash's theorem:

**Nash's Theorem.** *Let $(\mathcal{M}, g)$ be an $m$-dimensional Riemannian manifold and $f: \mathcal{M} \to \mathbb{R}^s$ a short smooth embedding (or immersion) into Euclidean space $\mathbb{R}^s$, where $s \geq m + 1$. This map is not required to be isometric. Then there is a sequence of continuously differentiable isometric embeddings (or immersions) $\mathcal{M} \to \mathbb{R}^s$ of $g$ which converge uniformly to $f$.*

Next, our work will take advantage of certain characteristics of AE(Autoencoder).

**Homomorphism of Autoencoders.** Given an AE, The discrete dataset denotes as $X$, and $x$ denotes mini-batch input, for $\forall x \in X$, there is:

$$f_d(f_e(x)) = x^r, \quad (2)$$

where the decoder $f_d: \mathcal{M} \to f_d(\mathcal{M})$, the encoder $f_e: \mathcal{N} \to f_e(\mathcal{N})$, and $x^r$ denotes the reconstruction. When the AE can losslessly reconstruct the original sample $x \cong x^r$, for $\forall x \in X$. The $f_d$ and $f_e$ are two smooth bijective mappings, thus, the $f_d$ could be regarded as a homeomorphic mapping.

**Immersion of the Decoder.** Based on the above description, let $x^z = f_e(x)$, in which the $h$th data in $x^z$ denotes as $x_h^z \in x^z$. If $f_d$ is a homeomorphic mapping, then the neighborhood of $x_h^z$, which denote as $U_{x_h^z}\mathcal{M}$, and the neighborhood after decoder mapping, which denote as $U_{f_d(x_h^z)}\mathcal{N}$, must be injective. According to the immersion definition, the $f_d$ can be regarded as an immersion from $U_{x_h^z}\mathcal{M}$ to $U_{f_d(x_h^z)}\mathcal{N}$.

## Tangent Space Approximation

In actual computer modeling, however, the construction of tangent space in high-dimensional Riemannian space is resource-intensive (Fontenele et al. 2014). Consequently, in this paper, we propose to use the neighborhood of a point on a manifold to approximately describe the tangent space expanded by this point. Take AE as an example: From the immersion properties of the decoder in the preliminary, it can be inferred that $f_d(x_h^z)|_U: U_{x_h^z}\mathcal{M} \to U_{f_d(x_h^z)}N$ could represent map transformation between neighborhoods on manifold, approximating the tangent mapping process $f_{d^*}(x_h^z): T_{x_h^z}\mathcal{M} \to T_{f_d(x_h^z)}\mathcal{N}$.

We propose leveraging the first-order Taylor expansion to analyze the above approximation map of the tangent space, the following equation describes the Taylor expansion approximation at point $f_d(x_h^z)$ on manifold $\mathcal{N}$:

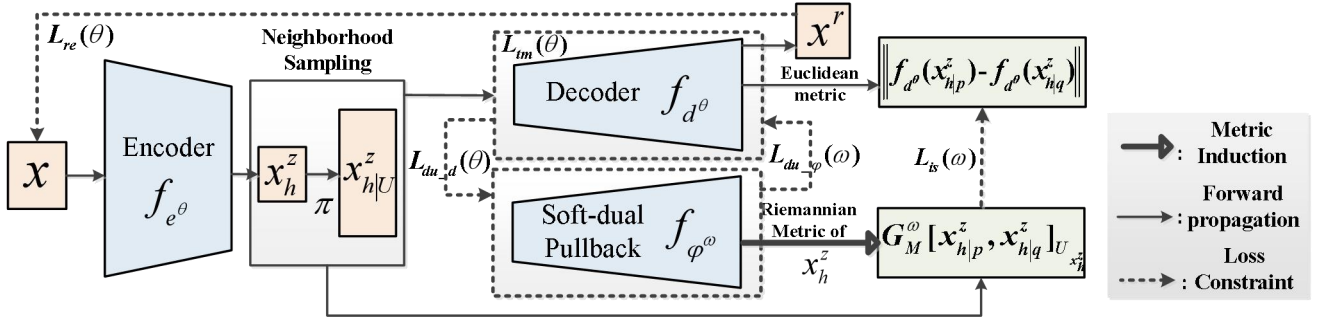

Figure 2: The structure of proposed model

$f_d(x^z_{h|p}) = f_d(x^z_h) + f_{d^*}(x^z_h) \cdot (x^z_h - x^z_{h|p}) + R(x^z_h; x^z_{h|p})$ (3)

where $f_d(x^z_{h|p}) \in U_{f_d(x^z_h)}\mathcal{N}$. The remainder $R$ is a high-order infinitesimal of $(x^z_h - x^z_{h|p})$, which denotes the error term for the approximation. Since $f_d(x^z_{h|p})$ is a point in the neighborhood $U_{f_d(x^z_h)}\mathcal{N}$, the smaller neighborhood $U_{f_d(x^z_h)}\mathcal{N}$ is, the smaller remainder $R$ will be. In this case, $f_d(x^z_{h|p})$ on the manifold can more accurately approximate the position $f_d(x^z_h) + f_{d^*}(x^z_h) \cdot (x^z_h - x^z_{h|p})$, which we denote as $f_{d^*}(x^z_{h|p})$ in the tangent space. Fig. 1 is utilized to visually illustrate the above approximate procedure.

## Proposed Method

Based on the theoretical framework in the preliminary, we propose the definition of isometric immersion learning:
**Isometric Immersion Learning** is to find a parameterized model that satisfies the immersion definition. Meanwhile, we expect this model to exhibit isometric properties during learning.

This section provides an overview of the model structure and training method we constructed based on the concept of isometric immersion learning. It is worth mentioning that the target manifold $\mathcal{N}$ of $f_d: \mathcal{M} \to \mathcal{N}$ in the Nash's theorem is a Euclidean space, so in the following sections, there is $\mathcal{N} = \mathbb{R}^s$. The Euclidean Normal $\|.\|$ is employed to calculate the distance.

### Isometric Immersion Model Structure

Firstly, we outline the insights of the structure.

As the immersion property can be achieved through the inherent homeomorphism of AE, it is logical to assume that the isometric definition can be satisfied by constraining the decoder $f_d$. To enhance the integration of the isometric notion into the model, we redefine isometric within the neighborhood. Let $f_{e^\theta}$ and $f_{d^\theta}$ be $\theta$-parameterized implementations of the ideal AE mentioned in Eq. (2). For $\forall x^z_h \in \mathcal{M}$, and $\forall x^z_{h|p}, x^z_{h|q} \in U_{x^z_h}\mathcal{M}$, we have:

$\mathcal{G}_\mathcal{M}[x^z_{h|p}, x^z_{h|q}]_{U_{x^z_h}} = \mathcal{G}_\mathcal{N}[f_{d^\theta}(x^z_{h|p}), f_{d^\theta}(x^z_{h|q})]_{U_{f_d(x^z_h)}}$ (4)

where $\mathcal{G}_\mathcal{M}[.,.]_{U_{x^z_h}}$ denotes the pullback metric within the neighborhood $U_{x^z_h}\mathcal{M}$.

Through theoretical deduction in the following, we prove that by employing certain training techniques, it could indeed ensure that $f_{d^\theta}$ satisfies immersion and induces a legal Riemannian metric. But, we also observe that it is impractical to demand $f_{d^\theta}$ maintain isometric while satisfying immersion. If do so, it may even reduce the probability of model convergence (the convergence probability observation experiment is included in the supplementary materials).

To address this, we propose a trick of separating the map $f_{d^\theta}$ and the pullback $f^*_{\varphi^\omega}$ with individual parameters $\theta$ and $\omega$. Thus, the Riemannian metric derived by $\omega$-parameterized pullback induction is $\mathcal{G}^\omega_\mathcal{M} = f^*_{\varphi^\omega}|_U \mathcal{G}_\mathcal{N}$. According to Definition 1, the $f^*_{\varphi^\omega}|_U$ should be the dual of the $f_{d^\theta}|_U$. Separate constrained tangent map and pullback weaken this dual bond and introduce soft dual loss, we refer to this trick as soft dual.

We exploit multiple loss functions to implement the techniques mentioned above. These loss functions allow us to trade-off between training convergence efficiency and theoretical rigor. The overall structure of the model is shown in Fig. 2. Next, we will analyze the loss functions of the new model.

**Immersion Loss.** To ensure that the AE follows the nearly homeomorphism, so that $f_{d^\theta}$ satisfies immersion. It is essential to have the output data to accurately reconstruct original input data:

$$\mathcal{L}_{re}(\theta) = \mathbb{E}_X[x - f_{d^\theta}(x^z)]^2 \quad (5)$$

**Tangent Space Approximation Loss.** The neighborhood on the manifold is employed to approximate the tangent space during the learning process. Based on Eq. (3), It is evident that a decrease in the size of the neighborhood leads to a reduction in the approximation error, which

| Algorithm 1: Isometric immersion learning |
|---|
| **Input**: Discrete dataset $X$. |
| **Parameter**: Initialize the parameters $\omega$ and $\theta$ in encoder $f_{e^\theta}$, decoder $f_{d^\theta}$, and the soft-dual pullback $f_{\varphi^\omega}$. Set a custom neighborhood sampling distribution $\pi$ and hyperparameter set $\{\alpha, \beta, \gamma, \varepsilon\}$. |
| **Output**: $f_{e^\theta}, f_{d^\theta}, f_{\varphi^\omega}$, through the differential of $f_{\varphi^\omega}$ to obtain the Riemannian metric $\mathcal{G}_{\mathcal{M}}^\omega$ on the manifold $\mathcal{M}$. |
| 1: **For** iteration < max iter number **do**: |
| 2:  Input the mini-batch $x$ into the encoder, and the latent code $x^z = f_{e^\theta}(x)$ was obtained. |
| 3:  Perform distribution sampling $\pi$ in the neighborhood of $x_h^z$ to obtain the sampling set $x_{h|U}^z$. |
| 4:   **For** iteration < max iter_imm number **do**: |
| 5:    Input $x_h^z$ and $x_{h|U}^z$ into the decoder |
| 6:    Calculate loss $\mathcal{L}_{immersion}(\theta) = \alpha \mathcal{L}_{re}(\theta) + \beta \mathcal{L}_{tm}(\theta) + \gamma \mathcal{L}_{du\_d}(\theta)$. |
| 7:    ADAM optimization of parameter $\theta$. |
| 8:   **For** iteration < max iter_iso number **do**: |
| 9:    The Riemann metric $\mathcal{G}_{\mathcal{M}}^\omega$ can be obtained by calculating $\mathcal{G}_{\mathcal{M}}^\omega = Df_{\varphi^\omega}^T \cdot Df_{\varphi^\omega}$. |
| 10:    Calculate loss $\mathcal{L}_{isometry}(\omega) = \varepsilon \mathcal{L}_{is}(\omega) + \gamma \mathcal{L}_{du\_\varphi}(\omega)$. |
| 11:    ADAM optimization of parameter $\omega$. |
| 12: **return** $f_{e^\theta}, f_{d^\theta}, f_{\varphi^\omega}, \mathcal{G}_{\mathcal{M}}^\omega$ |

motivates us to construct the following loss. Setting $\pi$ denotes a certain customized sampling distribution in the neighborhood $U_{x_h^z}\mathcal{M}$. The sample set $x_{h|U}^z \in U_{x_h^z}\mathcal{M}$ was obtained by performing $\pi$, and $\forall x_{h|p}^z, x_{h|q}^z \in x_{h|U}^z$. In order to reduce the error of approximate tangent space, the distance between sampling points within the neighborhood should be relatively small:

$$\mathcal{L}_{tm}(\theta) = \mathbb{E}_{X, x_{h|U}^z \sim \pi} \{\mathcal{G}_{\mathcal{N}}[f_{d^\theta}(x_{h|p}^z), f_{d^\theta}(x_{h|q}^z)]_{U_{f_d(x_h^z)}}\}$$
$$\xrightarrow{\mathcal{N}=\mathbb{R}^s} \mathbb{E}_{X, x_{h|U}^z \sim \pi} \left\| f_{d^\theta}(x_{h|p}^z) - f_{d^\theta}(x_{h|q}^z) \right\| \quad (6)$$

**Proof of Rieamannian Metric Legitimacy.** Due the potential of errors in the approximate learning process of parameterized networks, it is crucial to verify the legitimacy of the induced Riemannian metric: the metric matrix must be positive and symmetry. With mathematical deduction (in the supplementary material), we prove that if the immersion loss $\mathcal{L}_{re}$ is sufficiently low and the issue of neuron death in the final hidden layer of the encoder is avoided, the legitimacy of the Riemann metric could be guaranteed. Therefore, while building the network, we abstain from using ReLU activation functions that may result in neuronal death.

**Isometric Loss.** Consequently, inspired by Eq. (4), we build the following objective loss function to constrain the isometric property of the model:

$$\mathcal{L}_{is}(\omega)$$
$$= \mathbb{E}_{X, x_{h|U}^z \sim \pi}\{\mathcal{G}_{\mathcal{M}}^\omega[x_{h|p}^z, x_{h|q}^z]_{U_{x_h^z}} - \mathcal{G}_{\mathcal{N}}[f_{d^\theta}(x_{h|p}^z), f_{d^\theta}(x_{h|q}^z)]_{U_{f_d(x_h^z)}}\}^2$$
$$\xrightarrow{\mathcal{N}=\mathbb{R}^s} \mathbb{E}_{X, x_{h|U}^z \sim \pi}\{\mathcal{G}_{\mathcal{M}}^\omega[x_{h|p}^z, x_{h|q}^z]_{U_{x_h^z}} - \left\| f_{d^\theta}(x_{h|p}^z) - f_{d^\theta}(x_{h|q}^z) \right\|\}^2 \quad (7)$$

**Soft Dual Loss.** In actual training, in order for $f_{\varphi^\omega}^*|_U$ to become the dual mapping of the $f_{d^\theta}|_U$, we propose parameterizing $f_{\varphi^\omega}$, which should be nearly equal to $f_{d^\theta}$ as the dual mapping of $f_{\varphi^\omega}^*$. Therefore, since the Jacobian matrices of the two dual linear functions are transposed to each other, the Riemannian metric $\mathcal{G}_{\mathcal{M}}^\omega = Df_{\varphi^\omega}^T \cdot Df_{\varphi^\omega}$ in $\mathcal{L}_{is}$ could be calculated from $f_{\varphi^\omega}$'s Jacobian matrix $Df_{\varphi^\omega}$ during model training.

Therefore, for $f_{d^\theta}$, the loss function controlling dual softening is:

$$\mathcal{L}_{du\_d}(\theta) = \mathbb{E}_X(\| f_{d^\theta}(x^z) - f_{\varphi^\omega}(x^z) \|) \quad (8)$$

where the parameter $\omega$ in $f_{\varphi^\omega}$ is fixed, and the $f_{d^\theta}$ is optimized.

For $f_{\varphi^\omega}$, the loss function is:

$$\mathcal{L}_{du\_\varphi}(\omega) = \mathbb{E}_X(\| f_{d^\theta}(x^z) - f_{\varphi^\omega}(x^z) \|) \quad (9)$$

where we fix the parameter $\theta$ in $f_{d^\theta}$, optimize $f_{\varphi^\omega}$.

## Alternating Training Method

To train the new model with the previously offered multiple geometric priors. We propose leveraging the MLE method to optimize the parameters during model training. The optimization goal of isometric immersion learning is to find the MLE solution of the following objective by applying Nash's theorem with neighborhood approximation:

$$\text{argmax}_\omega \mathcal{P}(\mathcal{G}_{\mathcal{M}}^\dagger|_{U_{\mathcal{H}}}; \omega)$$

where $\mathcal{G}_{\mathcal{M}}^\dagger|_{U_{\mathcal{H}}}$ denotes the ideal pullback metric that makes $f_d|_U$ satisfy the isometric mapping in the neighborhood of $\forall h$, in which we set $\mathcal{H} = f_{e^\theta}(X)$, and $h \in \mathcal{H}$. Consequently, for $\forall h|_p, h|_q \in U_h\mathcal{M}$, we have:

$$\mathcal{G}_{\mathcal{M}}^\dagger[h|_p, h|_q]_{U_h} = \mathcal{G}_{\mathcal{N}}[f_d(h|_p), f_d(h|_q)]_{U_{f_d(h)}} \quad (10)$$

The loss function corresponding to this part is:

$$\mathcal{L}_{isometry}(\omega) = \varepsilon \mathcal{L}_{is}(\omega) + \gamma \mathcal{L}_{du\_\varphi}(\omega) \quad (11)$$

where $\varepsilon$, $\gamma$ denote the hyperparameters controlling the trade-off. When $\mathcal{G}_{\mathcal{M}}^\dagger$ holds, $\mathcal{L}_{isometry}$ should be nearly 0.

Moreover, The use of soft dual trick introduces a latent variable $\mathcal{Q}(f_d^\dagger(\mathcal{R})|_U; \theta)$, which denotes the ideal parameterized immersion $f_d^\dagger(r)|_U$ that can meet the immersion and induce a legitimate Riemannian metric in the neighborhood of $\forall r$, in which $\mathcal{R} = f_{e^\theta}(X)$, and $r \in \mathcal{R}$. The corresponding loss function is:

$$\mathcal{L}_{immersion}(\theta) = \alpha \mathcal{L}_{re}(\theta) + \beta \mathcal{L}_{tm}(\theta) + \gamma \mathcal{L}_{du\_d}(\theta) \quad (12)$$

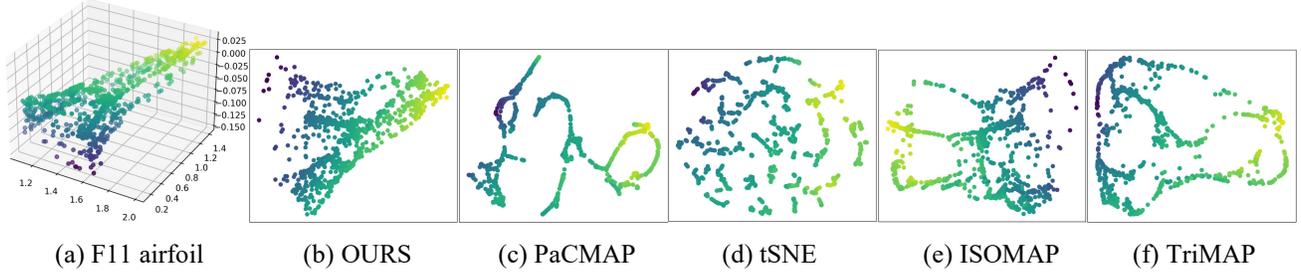

(a) F11 airfoil  (b) OURS  (c) PaCMAP  (d) tSNE  (e) ISOMAP  (f) TriMAP

Figure 3: Comparison of results of various manifold learning methods on F11 airfoil point cloud

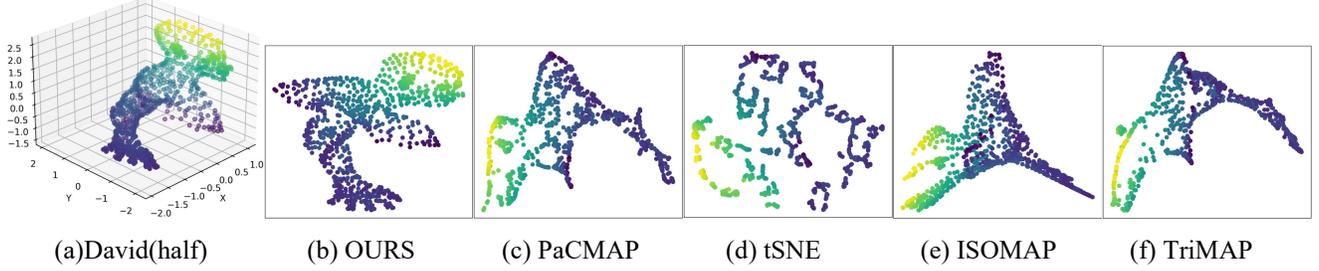

(a)David(half)  (b) OURS  (c) PaCMAP  (d) tSNE  (e) ISOMAP  (f) TriMAP

Figure 4: Comparison of results of various manifold learning methods on David (in TOSCA) point cloud

where $\alpha, \beta, \gamma$ denote the hyperparameters. When $f_d^\dagger$ holds, $\mathcal{L}_{immersion}$ should be nearly 0.

Therefore, $\mathcal{P}$ can be written as the logarithmic joint distribution form:

$$\ell(\mathcal{G}_M^\dagger|_{U_{\mathcal{H}}}; \omega) = \sum_h \ell n[\sum_r \mathcal{P}((\mathcal{G}_{\mathcal{M}}^\dagger|_{U_{x_h^z}}; \omega), (f_d^\dagger(r)|_U; \theta))] \quad (13)$$

We propose to leverage the EM algorithm to solve the Eq. (13). In addition, we also demonstrate the convergence of the model, whose comprehensive deduction is included in the supplementary material. The following briefly gives the results.

In step E:

$$\theta^{k+1} = argmax_\theta \mathcal{P}[(f_d^\dagger(\mathcal{R})|_U; \theta^k)|(\mathcal{G}_{\mathcal{M}}^\dagger|_{U_{\mathcal{H}}}; \omega^j)] \quad (14)$$

where the conditional probability $\mathcal{P}$ implies that we fix parameter $\omega^j$ of the $\mathcal{G}_M^\dagger$, and update the parameter $\theta^{k+1}$.

In step M:

$$\omega^{j+1} = argmax_\omega \sum_h \ell n \mathcal{P}[(\mathcal{G}_{\mathcal{M}}^\dagger|_{U_{x_h^z}}; \omega^j)|(f_d^\dagger(\mathcal{R})|_U; \theta^k)] \quad (15)$$

where, similarly, the conditional probability $\mathcal{P}$ implies that the parameter $\theta^k$ is fixed, and we update the parameter $\omega^{j+1}$ in the $\mathcal{G}_{\mathcal{M}}^\dagger$.

Intuitively, the entire network is updated alternately. The approximate process of immersion is optimized in step E to ensure that the learning process meets the geometry priors in $\mathcal{L}_{immersion}$. For the M step, under the premise that the immersion mapping is legal, a pullback $f_\varphi^{*\dagger}$ is sought so that the pullback metric $\mathcal{G}_{\mathcal{M}}^\dagger$ induced by it can ensure that $f_d^\dagger$ satisfies isometric. Based on above solution, alternating training method could be implemented as Algorithm 1.

## Experiments

**Datasets.** Due to the limitations of computility, we selected 3D surface manifold unstructured point cloud datasets as the experimental datasets, such as the TOSCA(Dutt et al. 2024) and the DLR-F11(Coder 2015). In the TOSCA dataset, we selected the point cloud structures of humans (David, Micheal, Victoria), cats, and gorillas as learning objects. We trimmed each point cloud set, extracted 800 to 2000 3D points of the object as the training set. In the DLR-F11 airfoil dataset, the shape of a half fuselage with an airfoil in the form of polynomial surface control points was provided. With the Bessel analytical expression of the wing shape, we obtained an unstructured point cloud training set consisting of 1,000 points on the airfoil based on random sampling of the analytical expression, as shown in Fig. 3 (a).

**Baselines.** We conducted manifold evaluation metric comparison experiments on the new model and various well-accepted manifold learning baselines, including: LLE(Roweis et al. 2000), MLLE(Zhang et al. 2006), LTSA(Zhang et al. 2004), t-SNE(Van der Maaten et al. 2008), UMAP(McInnes et al. 2018), TriMAP(Amid et al. 2019), Spectral(Ng et al. 2001), ISOMAP(Tenenbaum 2002), PaCMAP(Wang et al. 2021), and CAMEL(Liu 2024). The MLT-g(Hu et al. 2023) model and deep learning models of intelligent aerodynamic neighborhood such as DAN(Zuo et al. 2023) and MDF(Xiang et al. 2023) were used in the downstream tasks of the Riemannian metric learned by the new model.

| Type | | Global | | Local | | Both | Geometry | |
|---|---|---|---|---|---|---|---|---|
| | | Triplet | SpearCor | Npp | Nnwr | Auc | Curv_sim | Average |
| Local Methods | LLE | 0.772 | 0.759 | 0.325 | 0.337 | 0.394 | 0.711 | 0.550 |
| | MLLE | 0.851 | 0.862 | 0.505 | 0.592 | 0.557 | ***0.818*** | 0.698 |
| | LTSA | 0.844 | 0.792 | 0.007 | 0.001 | 0.001 | 0.531 | 0.363 |
| | tSNE | 0.794 | 0.733 | <u>***0.795***</u> | 0.966 | <u>***0.729***</u> | 0.434 | 0.742 |
| | UMAP | 0.745 | 0.718 | ***0.749*** | <u>***0.975***</u> | 0.621 | 0.023 | 0.639 |
| | TriMAP | ***0.915*** | ***0.923*** | 0.622 | 0.839 | 0.579 | 0.751 | ***0.772*** |
| Global Methods | Spectral | 0.805 | 0.831 | 0.498 | 0.603 | 0.495 | 0.703 | 0.656 |
| | ISOMAP | ***0.912*** | ***0.943*** | 0.588 | 0.759 | 0.611 | 0.737 | ***0.758*** |
| | PaCMAP | 0.847 | 0.832 | 0.688 | 0.891 | ***0.640*** | 0.193 | 0.682 |
| Geometric Methods | CAMEL | 0.811 | 0.752 | 0.596 | 0.813 | 0.554 | <u>***0.989***</u> | 0.753 |
| | **OURS** | <u>***0.934***</u> | <u>***0.964***</u> | ***0.713*** | ***0.908*** | <u>***0.757***</u> | ***0.948*** | <u>***0.871***</u> |

Table 1: Comparative experiments of manifold learning evaluations on the F11 dataset. We classify manifold learning models and metric evaluation methods by type. The best and second/third best results are emphasized in <u>***underlined***</u> and ***bold*** cases.

**Implementation Details.** In the Manifold Evaluation Metrics Experiments, we followed the same experimental settings as previous works (Liu 2024). In the downstream tasks of the Riemannian metric, we referred to the experimental setup in a recently published paper in the field of intelligent aerodynamics (Hu et al. 2023). All results are reported by taking the average over 10 trials. More detailed experimental settings are elaborated in the supplements.

## Manifold Evaluation Metrics

We compared the new model with 10 baselines under 6 recognized manifold evaluation metrics (Triplet(Amid et al. 2019), SpearCor(Huang et al. 2022), Npp(Huang et al. 2022), Nnwr(Liu 2024), Auc(Lee et al. 2015), Curv_sim(Liu 2024)). The Fig. 3 and Fig. 4 show the dimensionality reduction visualization results of the new method and other methods on two datasets. Table 1 provides the manifold learning evaluation results on the DLR-F11 point cloud, and the evaluation metrics scores are all between 0 and 1. The closer to 1, the better the evaluation result. We marked the top three best results of each metric and calculated the average of all evaluation metrics. From the experimental results, it can be seen that the new model can achieve outstanding results under most evaluation metrics. More manifold evaluation metrics studies of other datasets are in the supplements. It is worth noting that, the new model has the highest average metric score compared with the other 10 baselines in all datasets.

In addition, we conducted a comparison of the isometric mapping distortion evaluation on various datasets. We quantify distortion values with $\mathcal{L}_{is}$. Traditional manifold learning models do not have the ability to conduct metric learning. As a result, the metric on their learned manifold is the Euclidean metric $l_2$ throughout. We concretized $\pi$ for these methods using the K-nearest neighbors (KNN) algorithm. From the results shown in Table 2, it is evident that the new model consistently exhibits an apparently lower isometric distortion compared to the other models in all datasets. These results demonstrated the effectiveness of the new model under different metrics.

| | David | Cat | Micheal | DLR-F11 | Swiss_Roll |
|---|---|---|---|---|---|
| LLE | 4.951 | 6.253 | 6.109 | 5.567 | 0.034 |
| MLLE | 4.954 | 6.258 | 6.138 | 5.573 | 0.034 |
| LTSA | 4.952 | 6.259 | 6.139 | 5.567 | 0.035 |
| ISOMAP | 0.695 | 0.869 | 2.073 | 1.278 | 0.007 |
| tSNE | 30.81 | 40.26 | 68.06 | 16.11 | 55.01 |
| Spectral | 4.962 | 6.264 | 6.152 | 5.582 | 0.035 |
| CAMEL | 2.637 | 4.696 | 4.198 | 3.984 | 5.068 |
| PaCMAP | 3.453 | 4.612 | 5.271 | 4.647 | 3.769 |
| UMAP | 3.751 | 5.050 | 4.843 | 4.816 | 0.807 |
| TriMAP | 6.920 | 4.918 | 12.75 | 6.070 | 48.32 |
| **OURS** | <u>**0.023**</u> | <u>**0.100**</u> | <u>**0.630**</u> | <u>**0.858**</u> | <u>**0.002**</u> |

Table 2: The isometric mapping distortion of various manifold learning methods (David, Cat and Micheal belong to TOSCA dataset). The Minimum distortion are emphasized in <u>**underlined**</u> cases.

## Downstream Tasks of the Learned Metric

The new model provides the capacity to learn Riemannian metrics. However, the Riemann metric is a tensor that does not possess constant components that can be compared. Therefore, we tested the practicality of the metrics learned by the new model based on the performance of downstream tasks. Specifically, we observed the impact of

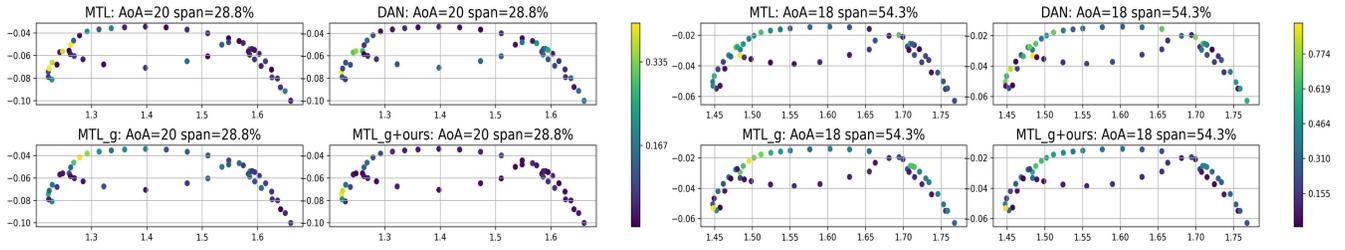

(a) Slice at 28.8% on F11 airfoil with AoA of 20°

(b) Slice at 54.3% on F11 airfoil with AoA of 18°

Figure 5: Comparison of absolute error of pressure coefficient prediction at the sampling points of the airfoil slice. The darker the color of point, the smaller the prediction error

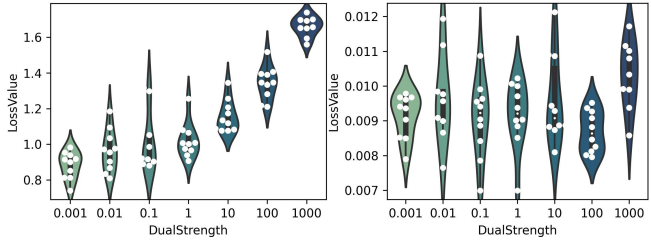

(a) Impact on $\mathcal{L}_{is}$

(b) Impact on $\mathcal{L}_{re}$

Figure 6: The impact of dual ablation on model's learning objectives

| AoA | MTL | MDF | DAN | MTL_g | MTL_g+ ours | Δ |
|---|---|---|---|---|---|---|
| 7° | 1.493 | 1.532 | 1.518 | 1.464 | **1.443** | 1.4% |
| 12° | 1.119 | 1.214 | 1.213 | 1.225 | **1.142** | 6.7% |
| 16° | 0.826 | 0.666 | 0.654 | 0.654 | **0.581** | 11.0% |
| 18° | 0.331 | 0.292 | 0.228 | 0.202 | **0.181** | 10.4% |
| 18.5° | 0.222 | 0.182 | 0.156 | 0.149 | **0.123** | 17.2% |
| 19° | 0.171 | 0.208 | 0.079 | 0.077 | **0.071** | 7.0% |
| 20° | 0.170 | 0.069 | 0.023 | 0.016 | **0.014** | 7.7% |
| Avg | 0.619 | 0.595 | 0.553 | 0.541 | **0.508** | 8.8% |

Table 3: The average MSE error of the aerodynamic pressure coefficient prediction task compared with various intelligent aerodynamics model. Where Δ represents the improvement percent of the learning accuracy of the learned Riemann metric model compared with the original MTL_g. AoA: Angle of Angle of attack. The best results in each AoA are emphasized in **underlined** cases.

the Riemannian features learned by the new model and the mathematically derived Riemannian features on the results of the same learning task.

We referred to the work on intelligent aerodynamics in Riemannian space, which utilized Riemannian features to predict aerodynamic coefficients on the F11 airfoil (Hu et al. 2023).

In the original work, the mathematical analytical expression of Bezier surfaces was leveraged to calculate the Riemannian manifold features such as Riemannian metric, connection, and curvature on the airfoil surface.

In contrast, our proposed model can obtain Riemannian features by unsupervised learning the geometric structure of the F11 airfoil's point clouds.

The features obtained by the above two methods are input into the same MLT_g network proposed in the original work, and the pressure coefficient on the wing is predicted in a supervised manner. The prediction error comparison results are shown in Table 3 and visualized in Fig. 5. The results indicate that, within the identical learning framework, the model that uses learned Riemann features outperforms the model that utilizes Riemann features derived from the Bezier surface in terms of prediction accuracy. The improved accuracy of the predictions implicitly validates the accuracy and practicality of the metrics obtained by the new model.

**Soft Dual Ablation Studies**

We conducted ablation studies in terms of soft dual. The results with the DLR-F11 dataset are shown in Fig. 6. The correlation between the strength of the dual bond and the isometric loss of the model is clearly apparent. The weaker the dual bond, the lower the isometric loss $\mathcal{L}_{is}$. Relatively, soft dual has a smaller impact on immersion loss $\mathcal{L}_{re}$. It is essential to mention that we also performed the same experiment on models that did not employ the soft dual trick. The $\mathcal{L}_{is}$ and $\mathcal{L}_{re}$ of these models were around 10 times higher on average compared to those that utilized soft dual trick.

## Conclusion

In this paper, we propose a neural network-based model to implement isometric immersion learning in Riemannian space. We theoretically derive and algorithmically implement an alternating training method that aligns with the isometric immersion concept. The effectiveness and practicality of our new model for learning isometric manifolds have been demonstrated via experiments.